\documentclass[conference]{IEEEtran}
\IEEEoverridecommandlockouts
\usepackage{cite}
\usepackage{amsmath,amssymb,amsfonts}
\usepackage{algorithmic}
\usepackage{graphicx}
\usepackage{textcomp}
\usepackage{xcolor}
\usepackage{multirow}
\usepackage{booktabs}
\usepackage{url}
\def\BibTeX{{\rm B\kern-.05em{\sc i\kern-.025em b}\kern-.08em
    T\kern-.1667em\lower.7ex\hbox{E}\kern-.125emX}}
\usepackage{hyperref}
\hypersetup{
    colorlinks=true,
    linkcolor=blue,
    filecolor=magenta,      
    urlcolor=blue,
    pdftitle={Overleaf Example},
    pdfpagemode=FullScreen,
}
\begin{document}

\title{
% Comparing unstructured knowledge injection methods for LLMs\\
Comparing Knowledge Injection Methods for LLMs in a Low-Resource Regime
% Comparing augmentation strategies for unstructured knowledge injection for LLMs
% Comparing augmentation methods for unstructured knowledge injection in LLMs
% \thanks{Identify applicable funding agency here. If none, delete this.}
}

\author{%
\IEEEauthorblockN{Hugo Abonizio$^{1,4}$,
Thales Almeida$^{2,4}$,
Roberto Lotufo$^{1,3}$,
Rodrigo Nogueira$^{4}$}
\IEEEauthorblockA{$^{1}$Faculdade de Engenharia Elétrica e de Computação (FEEC), University of Campinas (Unicamp)}
\IEEEauthorblockA{$^{2}$Instituto de Computação (IC), University of Campinas (Unicamp)}
\IEEEauthorblockA{$^{3}$NeuralMind}
\IEEEauthorblockA{$^{4}$Maritaca AI, Campinas, SP -- Brazil}
}

\maketitle

\begin{abstract}
Large language models (LLMs) often require vast amounts of text to effectively acquire new knowledge. While continuing pre-training on large corpora or employing retrieval-augmented generation (RAG) has proven successful, updating an LLM with only a few thousand or million tokens remains challenging. In this work, we investigate the task of injecting small, unstructured information into LLMs and its relation to the catastrophic forgetting phenomenon.
We use a dataset of recent news -- ensuring no overlap with the model's pre-training data -- to evaluate the knowledge acquisition by probing the model with question-answer pairs related the learned information.
Starting from a continued pre-training baseline, we explored different augmentation algorithms to generate synthetic data to improve the knowledge acquisition capabilities.
Our experiments show that simply continuing pre-training on limited data yields modest improvements, whereas exposing the model to diverse textual variations significantly improves the learning of new facts -- particularly with methods that induce greater variability through diverse prompting.
Furthermore, we shed light on the forgetting phenomenon in small-data regimes, illustrating the delicate balance between learning new content and retaining existing capabilities.
We also confirm the sensitivity of RAG-based approaches for knowledge injection, which often lead to greater degradation on control datasets compared to parametric methods.
Finally, we demonstrate that models can generate effective synthetic training data themselves, suggesting a pathway toward self-improving model updates.
All code and generated data used in our experiments are publicly available, providing a resource for studying efficient knowledge injection in LLMs with limited data at \url{https://github.com/hugoabonizio/knowledge-injection-methods}.
% https://anonymous.4open.science/r/ijcnn2025/.

% We propose a method to measure the effectiveness techniques for injecting free-form knowledge into LLMs. Previous work use entity-centric evaluation methods in which the piece of information to be learned can be expressed as triples in a knowledge graph and rule-based metrics to measure the injection effectiveness. However, not all real-world information can be fully expressed as triples. We argue that those evaluation methods fail to capture whether an LLM learned that information. We show that allowing LLMs to generate free-form answers can trully expose the knowledge learned.
\end{abstract}

\begin{IEEEkeywords}
Large language models, Knowledge injection, Data augmentation, Synthetic data
\end{IEEEkeywords}

\section{Introduction}

Previous work has shown that large language models (LLM) trained with a self-supervised objective learn large amounts of information, effectively acting as knowledge bases~\cite{petroni2019language,heinzerling-inui-2021-language,wang-etal-2021-generative,zhong-etal-2021-factual,youssef-etal-2023-give}.
Likewise, updating the knowledge of a model through continued pre-training~\cite{gururangan-etal-2020-dont,Tongtong2024}, with the goal of specializing an existing LLM in a domain such as mathematics~\cite{lewkowycz2022solving,azerbayev2024llemma}, medicine~\cite{singhal2023large,xie2024llama} or
code~\cite{rozière2024codellamaopenfoundation}, has also shown to be fruitful. In these scenarios, the amount of training data spans from billions to trillions of tokens.

However, perhaps surprisingly, incorporating relatively small amounts of information (e.g., thousands or millions of tokens) has proven to be more challenging, often resulting in performance degradation or only marginal gains when done naively~\cite{ciosici-etal-2021-perhaps,ke2023continual,ovadia2023fine,cheng2024adapting,biderman2024lora}, and potentially even increasing \textit{hallucinations}~\cite{gekhman-etal-2024-fine}.
An illustrative case is that, while one might expect fine-tuning on new data using self-supervision to enable the LLM to internalize additional information into its parametric knowledge, it has not been shown to seamlessly learn certain relational inferences (e.g., from ``A is B'' to ``B is A'')~\cite{berglund2024the}.
% For instance, updating the knowledge of an LLM so it learns a the takeaways of a single scientific article is not seamlessly, as simply pre-training in self-supervised manner on a document does not guarantee that the model will internalize that piece of information into its parametric knowledge. The literature shows examples of this difficulty 

An additional complication is the forgetting problem, in which new information can often be successfully injected at the expense of forgetting previously learned knowledge~\cite{Kirkpatrick2017,riemer2018learning,kleiman2023predicting,Luo2023AnES,biderman2024lora,kotha2024understanding}, a phenomenon commonly referred to as catastrophic forgetting~\cite{MCCLOSKEY1989109,FRENCH1999128,Goodfellow2014}.

To address these problems, recent literature on knowledge injection often concentrates on two extremes. In the first, also referred to as model editing, methods that require the information to be learned can be expressed as well-defined entities and relations~\cite{zhu2020modifyingmemoriestransformermodels,meng2022locating,meng2023massediting,berglund2024the}. However, the techniques proposed in these works cannot be easily applied to real documents without modifications, as it is challenging to convey the complex information as a knowledge graph with a finite number of possible relationships.
% For instance, we challenge the reader to convert to the triples the following the news headline:\\
For example, converting the following headline into a set of discrete triples presents a non-trivial challenge:\\
``\textit{Advocates for Ukraine’s surprise incursion into the Russian territory say it will provide Kyiv with vital leverage for any future peace talks (December 2nd, 2024)}''

On the other end of the spectrum, works on domain adaptation methods often builds upon datasets on the order of billions of tokens to learn new information~\cite{rozière2024codellamaopenfoundation,cheng2024adapting,cheng-etal-2024-instruction,ibrahim2024simple}. These approaches require substantial computational resources and are, therefore, only feasible for organizations with large-scale infrastructure and access to vast amounts of data. In many private applications or niche domains, however, data can be scarce, and compute resources are limited, making such large-scale methods impractical in some real-world settings.

In parallel, retrieval-augmented generation (RAG)~\cite{NEURIPS2020_6b493230} methods aim to inject knowledge through in-context learning, rather than by updating the model's parametric knowledge. Some studies have compared these two approaches, highlighting different advantages in each case~\cite{balaguer2024ragvsfinetuningpipelines,mecklenburg2024injectingnewknowledgelarge}. However, it is important to note that these approaches are orthogonal: parametric knowledge injection and in-context learning methods can be combined and are not mutually exclusive.

In this work, we focus on the middle ground between the two extremes. Our goal is to study the learning dynamics of small, unstructured information in an efficient manner, without forgetting previously acquired knowledge. We investigate the learning–forgetting tradeoff using a small corpus and evaluate different continual pre-training techniques including augmentation algorithms that leverage synthetic variations of the original data, aiming to overcome the challenges of learning in a small-data regime.

To measure the learning effectiveness, we need a dataset containing information that is both new to the model -- i.e., it was not seen during its pre-training -- and complex, to make the results impactful for real-world applications. Thus, we chose the TiEBe dataset~\cite{almeida2025tiebebenchmarkassessingcurrent}, which contains news articles with the required recency, and question-answer pairs to measure knowledge and understanding.

% To measure the learning effectiveness, we collected a set of news documents published between 2015 and 2024 and used questions generated specifically to probe the knowledge about these facts. To ensure the novelty of the learned information, we evaluated the models on the most recent documents, published and describing events that happened after the release of the model. This dataset took inpiration from prior work that proposed recent or realtime events dataset~\cite{temporalwiki,vu2023freshllms,kasai2023realtime,ovadia2023fine,wu2024finetunebenchcommercialfinetuningapis,jiang-etal-2024-instruction}. However, those datasets lack the required recency to avoid possible contamination, i.e., informations already present in the training data of the model. Thus, we created a new collection to fill the gap of having documents as recent as the end of 2024 to evaluate current models while mitigating the risk of contamination. All data used in the experiments, including the evaluations questions, training datasets and their augmentations are publicly available\footnote{https://[URL]}.

Our results indicate that directly continuing pre-training on a set of documents using self-supervised learning~\cite{gururangan-etal-2020-dont} (i.e., next-token prediction) has limited effectiveness. Additionally, augmentation methods suggest that training on multiple versions of the same source document is necessary. Our findings also shed light on why LLMs require vast amounts of training data: intuitively, learning a small piece of information should not require 20 variations -- one or two should suffice. Addressing this inefficiency could make training these models -- which currently cost millions of dollars -- significantly more affordable.

In summary, our main contributions are as follows:

\begin{itemize}
% \item We release a novel dataset to probe factual knowledge regarding recent events that most models release to date have not been exposed to.

% adicionar a metodologia de avaliação como contribuição
% highlight que a contribuição é o método de avaliação de injestão de conhecimento
% os augmented data with reproduction code and detailed results and outputs também são uma contribuição
% \item Additionally, release a set of questions, answers, and the context that grounds them to automatically evaluate the learning of new information.

\item We conduct an extensive analysis of different knowledge injection approaches, including RAG, continued pre-training and augmentation techniques. We compare different techniques and their variations, providing evidence that models benefit from exposure to diverse data variations to effectively learn new information.
\item We propose an evaluation methodology to assess the effectiveness of knowledge injection using small and unstructured datasets.
\item We provide insights into the challenges of continued pre-training in small-scale data regimes and propose strategies to address training instabilities in this context.
\item We release a set of synthetically augmented corpora along with the code to reproduce and expand them, supporting future research.
% falar da disponibilização do código e dos textos augmentados (citar o número de variações)
\end{itemize}

\section{Related Work}

Recent work on injecting new knowledge into LLMs can be divided into three categories: (1) model editing techniques that modify entity relationships within the model parameters, (2) knowledge injection through in-context learning, and (3) knowledge injection via some form of continual training. In this paper, we turn our attention to the latter two, examining the dichotomy between retrieval-augmented generation and fine-tuning.

% Orthogonal to our topic of interest is machine unlearn~\cite{} whose goal is to forget specific knowledege such as a person's private information that was once learned by the model.
% Once again, the majority of these 
The majority of model editing works focus on entity-centric tasks~\cite{eldan2023whosharrypotterapproximate}. For instance Zhang et al.~\cite{zhang2024comprehensivestudyknowledgeediting} propose KnowEdit, a benchmark consisting mostly of entity-centric datasets, such as Wikidata~\cite{hartvigsen2023aging}, ZsRE\cite{levy2017zero}, and WikiBio~\cite{cohen2024evaluating}.
Eva-KELLM~\cite{wu2023evakellmnewbenchmarkevaluating} is another example of benchmark, in which the source of information to be learned comes from a raw document instead of a triple. However, in their evaluation methodology they still use triples to evaluate whether the knowledge was successfully learned.
 In our work, we depart from this constraint and evaluate the effectiveness of knowledege injection techniques by letting the LLM generate free-form answer for a given prompt that probes wheter the model knows a particular piece of information.
 
A common approach to knowledge ingestion involves coupling the LLM with a retriever that has access to a database containing the relevant information~\cite{zhang2024synthetic,gao2024retrievalaugmentedgenerationlargelanguage}, a method generally referred to as RAG~\cite{NEURIPS2020_6b493230}. Because it does not require model training, it is relatively cost-effective for real-world scenarios. However, later in this work, we will demonstrate that this approach has some drawbacks. Some of the most frequently discussed in the literature include its dependence on retriever quality~\cite{finardi2024chroniclesragretrieverchunk}, the chunking strategy~\cite{zhong2024mixofgranularityoptimizechunkinggranularity}, and the model’s ability to handle the provided context~\cite{liu-etal-2024-lost}.

% We will later show that selecting the retrieval collection is not a easy task as improvements in one domain led to degration in others when RAG turned on, and automaticaly deciding when to search is not a trivial task for LLMs, especially smaller ones.

Finally, in the last category of related work are methods that leaveraged continued training, either by continual pre-training or fine-tuning. Here, we refer to continual pre-training as the self-supervised training using next-token prediction, while fine-tuning methods are the ones trained using instruction-tuning~\cite{wei2022finetuned}.

Ovadia et al.~\cite{ovadia2023fine} compared RAG with continual pre-training on synthetic paraphrases, finding that exposing the new information in diverse ways through paraphrases play an important role. Their results found that RAG knowledge injection to be more effective than self-supervised finetuning. The work also generated questions to probe the effectiveness of knowledge injection.

Cheng et al.~\cite{cheng2024adapting,cheng-etal-2024-instruction} proposed methods for synthetically augmenting the pre-training corpus by transforming the original examples using different tasks. Their results show advantages over vanilla pre-training in the original documents, highlighting the importance of variations for effective learning.

Balaguer et al.~\cite{balaguer2024ragvsfinetuningpipelines} and Mecklenburg et al.~\cite{mecklenburg2024injectingnewknowledgelarge} compared RAG and fine-tuning methods by training on synthetic pairs of questions and answers. Both work showed a significant increase in performance after fine-tuning and Balaguer et al. showed that RAG and fine-tuning can be combined synergistically.

Wu et al.\cite{wu2024finetunebenchcommercialfinetuningapis} also investigated the knowledge injection through fine-tuning on question-answer (QA) pairs and used recent news as one of their evaluation datasets. Their results showed that learning from this fine-tuning is limited.

Yang et al.~\cite{yang2024syntheticcontinuedpretraining} introduced the EntiGraph, a knowledge graph-based augmentation technique for generating diverse synthetic text. They continued pretraining on small, domain-specific corpora and showed that their approach outperformed standard continued pretraining and simple paraphrasing using question-answering accuracy. Additionally, they found that their approach is complementary to RAG, improving downstream performance even when the original documents were available at inference time. While their work is closely related to ours, we place a stronger emphasis on mitigating data contamination risks and explore knowledge injection with an even smaller training corpus.

% We add a nuanced discussion to those findings by showing that choosing the correct retrieval corpus is not trivial, and that RAG methods detiriorate LLM performance on unrelated tasks.

% shows the risks of injecting new knowledge through fine-tuning a pre-trained model. Their results conclude that it may lead the models to \textit{hallucinate} more. 

\section{Methods}

In this section, we describe our evaluation methodology, including the dataset used and the knowledge injection techniques evaluated.

\subsection{Dataset}

We chose to inject knowledge related to recent news articles for two reasons. First, news articles carry complex forms of knowledge expression, which we argue are more aligned with real-world challenges researchers and practitioners will face when keeping an LLM continuously up-to-date. This contrasts with simpler forms of knowledge, such as facts encoded as knowledge triples (e.g., ``John Smith works at ACME Corporation''). A news article might be incomplete (e.g., it mentions key people participating in an ongoing event but does not define their roles, assuming the reader has been following the event for a while), or it might contradict other documents (e.g., ``investment X is no longer recommended due to fraud scandals'').

Second, using the news domain we can ensure that the model has not been exposed to that specific information previously. Given the recency of the news, we can mitigate the contamination problem using news about events that occurred after the model's training cutoff.

To address these two criteria, we used the TiEBe dataset~\cite{almeida2025tiebebenchmarkassessingcurrent}\footnote{\url{https://huggingface.co/datasets/TimelyEventsBenchmark/TiEBe}}, a dataset of news articles spanning from 2015 to 2024, along with a corresponding set of question-answer pairs, to evaluate models on their knowledge of specific events. The questions and answers were generated using \texttt{GPT-4o-2024-08-06}, by prompting it to produce four pairs per article in a single generation. More details on the creation of the data set can be found in Almeida et al.~\cite{almeida2025tiebebenchmarkassessingcurrent}.

To run the knowledge injection experiments, we chose the Llama-2 model~\cite{touvron2023llama2openfoundation}, which has an old enough knowledge cutoff -- September 2022 -- while having strong performance in the question-answering task due to being instruction-finetuned.
This mitigates the risk of data contamination because the model was not exposed to the specific events covered in the documents during the pre-training.

However, since Llama-2 has a limit of 4k tokens in its context, we selected a subset of the \textit{World} category filtering articles with up to 3,500 tokens to fit the article, the instruction template, and the generated answer within the model's context length.
Additionally, we selected only the recent documents, from 2023 to 2024, as the training corpus. The final dataset comprises 117 documents, each paired with four QA pairs -- 468 QA pairs, in total.

For the automatic evaluation of the models' answers, we followed the methodology described in Almeida et al., applying the process commonly known as LLM-as-a-judge~\cite{zheng2023judging,gu2025surveyllmasajudge}.
This approach leverages the expected answers provided in the dataset, prompting an LLM to assess the correctness of candidate answers based on these true answers.
This more sophisticated way of evaluating the answers is required to check whether the model has learned the complex facts contained in the documents, whereas strict approaches, such as exact matching, may fail to capture these nuances.
The prompt and evaluation code we used are available in the released repository.

\subsection{Control Datasets}
One of the challenges in updating a model's knowledge is ensuring it retains previously learned capabilities. Large models can easily memorize (i.e., fully reconstruct) a small set of documents, which is a key goal of the knowledge injection task. However, this often comes at the cost of significant performance drops on unrelated tasks where the original model previously excelled.

To quantify this forgetting gap, we use the average accuracy across the following seven datasets, collectively referred to as the \textit{Control datasets}: OpenBookQA~\cite{OpenBookQA2018}, ARC-Easy and ARC-Challenge~\cite{allenai:arc}, WinoGrande~\cite{sakaguchi2019winogrande}, HellaSwag~\cite{zellers-etal-2019-hellaswag}, PIQA~\cite{Bisk2020}, and BoolQ~\cite{clark-etal-2019-boolq}. These datasets are implemented in the Language Model Evaluation Harness~\cite{eval-harness}, which has been used in prior work for similar evaluations~\cite{ma2023llmpruner,zhang2024tinyllama,eldan2024whos}.

\subsection{Knowledge Injection Techniques}
\label{sec:injection_techniques}
We investigate the following knowledge injection techniques:

\textbf{Retrieval-Augmented Generation (RAG)}: One of the simplest and most straightforward methods for injecting knowledge into an LLM is to use a retrieval mechanism to locate relevant information within a corpus and include this information in the prompt provided to the LLM. Specifically, we employ BM25~\cite{Robertson1994OkapiAT,rosa2021yesbm25strongbaseline} to retrieve the top-N documents from the corpus of recent news, which consists of 117 documents.

To evaluate different configurations, we tested a document retrieval approach, where the best-matching document is prepended to the prompt, followed by the test question. Additionally, we evaluated a chunking-based approach, where documents are divided into chunks of 512 tokens with an overlap of 64 tokens. In this setup, we retrieved the top-5 chunks, which were then prepended to the question to allow the LLM to generate an answer.

\textbf{Continual Pre-training (CPT)}: This approach involves continuing the pre-training of the LLM directly on the target document or article using the causal language modeling objective (i.e., next-token prediction) applied to the unmodified corpus~\cite{gururangan-etal-2020-dont}.

\textbf{Rephrasing the Web (RTW)}: Following the four prompts proposed by Maini et al.~\cite{maini-etal-2024-rephrasing}, we generate rephrased versions of a training example. Three of these prompts instruct the LLM to rephrase the input document in styles with varying levels of complexity: easy (simplified, suitable for a toddler), medium (clear and high-quality, similar to Wikipedia), and hard (terse and abstruse). The fourth prompt asks the LLM to generate QA pairs that the document is likely to address.

We experiment with two models for generating these rephrases: (1) GPT-4o,\footnote{more specifically, GPT-4o-2024-08-06} which provides a high-quality upper bound for rephrases but may represent an unrealistic setup since it could have been exposed to the content during its pre-training;\footnote{\url{https://platform.openai.com/docs/models}} and (2) the model itself (i.e., Llama-2-7B), representing a more practical scenario in which the model does not have prior knowledge of the document being rephrased.

We report the results using all the four styles of prompts, using only the first three, and using only the QA-style prompt. This way we can keep only the rephrasing prompts and isolate the possibility of cross-contamination by generating similar questions as the ones used during the test phase.

% To ensure that the model never have seen the news articles, we use only news from 2023 and on. Since llama-2-7b knowledege cutoff date is September 2022, this garantees the model has not seens this data.

% Since llama-2-7b is not instruction-tuned, we use a N-shot prompt to guide the model into generating answers in the proper format.

% Once the model is trained with one of the techniques described above, we use a evaluation question as input to the model, sample an answer (temp=0?), and use GPT-4o-2024-08-06 as an LLM-as-a-judge to evaluate whether the generated answer matches the ground-truth one.\footnote{The prompt for the LLM as a judge can be found in Appendix A.2}. The judge outputs a binary metric: whether the answer is correct or not. Notice that in this evaluation phase we don't use the news article as input to the model; it needs to generate answer from memory.

\textbf{Instruction Pre-training (IPT)}: We applied the instruction generation method proposed by Cheng et al.~\cite{cheng-etal-2024-instruction} to our training corpus. Using their instruction generation model,\footnote{\url{https://huggingface.co/instruction-pretrain/instruction-synthesizer}} we generated synthetic instructions for each training document in a 1-shot setting. A 1-shot approach was necessary, as using more examples would result in truncation due to exceeding the model's context length.

\textbf{Paraphrasing (Para)}: To evaluate the effect of simple paraphrasing on training examples, we adapted the prompt proposed by Ovadia et al.~\cite{ovadia2023fine}. The prompt instructed the LLM to rephrase the content while maintaining factual accuracy and maintaining the original text length. We generated multiple paraphrases by applying token sampling with a specified temperature. For this process, we used two models: GPT-4o, providing high-quality paraphrases as an upper bound, and Llama-2-7B, representing a more realistic, self-contained approach.

\subsection{Training Setup}

The knowledge injection methods relying on continued pre-training were implemented by mixing synthetic augmented examples with the original documents. Therefore, all reported results, along with the corresponding variations, refer to the original 117 documents supplemented with an additional N synthetic variations in the training set.

Starting from the Llama-2-7B-chat checkpoint\footnote{\url{https://huggingface.co/meta-llama/Llama-2-7b-chat-hf}}, we further trained the model using the causal language modeling objective with the traditional cross-entropy loss. Training was conducted on batches of 8 examples, with a learning rate of \texttt{5e-5}, and the AdamW~\cite{loshchilov2018decoupled} optimizer. Given the task of injecting small amounts of knowledge, our training runs were intentionally short, often involving fewer than 15 training steps per epoch. To ensure stable training and reduce variance in results, we carefully tuned the hyperparameters in preliminary experiments, determining that training for two epochs with a relatively large learning rate warmup yielded the best performance.

As described in \cite{ibrahim2024simple}, we applied a re-warmup and re-decay strategy to the learning rate, utilizing linear warmup and cosine decay. Specifically, the warmup phase was applied throughout the first epoch, while the decay phase occurred during the second and final epoch. This approach allowed us to re-warm the learning rate during the first half of training and re-decay it during the second half, optimizing the stability and effectiveness of the training process.

%However, given our smaller data regime and preliminary experiments showing more stable training with this approach, we applied the warmup throughout the first epoch and the decay during the second -- and final -- epoch. This way, we re-warmed up the learning rate for the first half of the training and re-decayed it for the remaining half.

% Because of this, we found that small modifications in the training hyperparameters, might lead to high variations in the final results. After some various trial, we found that using a unusual large learning rate warmup stage that spans the entire first epoch, followed by a cosine decay in the second epoch, highly decrease the variations.

\section{Results}

In this section, we describe and analyze the results of our experiments and conduct ablations comparing variations of the studied methods.

% \subsection{How many samples are necessary?}
\subsection{Which method is the best?}

\begin{figure}
    \centering
    \includegraphics[width=\linewidth]{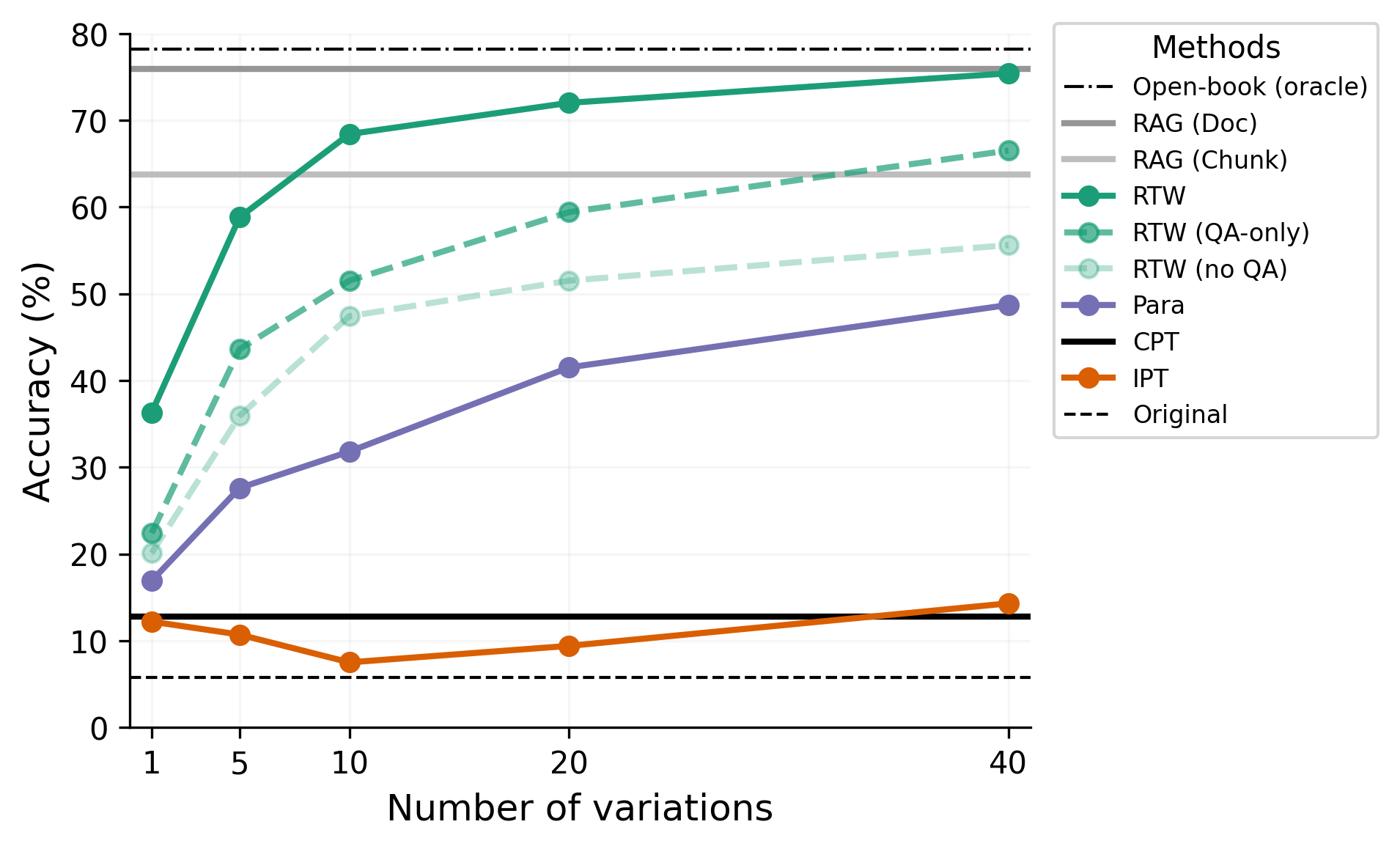}
    \caption{Comparison of different knowledge injection methods, including parametric and non-parametric techniques. The upper bound is represented by open-book answering with access to the source document (oracle), while the lower bound corresponds to the model's original performance in closed-book answering. Colored lines represent the knowledge injection methods evaluated in this study. The y-axis represents accuracy on the TiEBe dataset, considering only events from 2023 and 2024. The x-axis indicates the number of variations used for each augmentation method.}
    \label{fig:variations}
\end{figure}

Fig.~\ref{fig:variations} summarizes the comparison of the methods evaluated in this work. The dashed black lines show the lower and upper bounds, which correspond to the closed-book answering performance (when the model relies solely on its parametric knowledge) and the open-book answering performance (when the model has access to the context that answers the given question) using the original model.

Next, still using open-book answering but without providing the exact correct context (oracle), we evaluated RAG approaches. This is a more realistic scenario because in real-world applications the pairing of a question and its relevant context is not known a priori. For RAG evaluation, we used the top-1 most similar document according to the BM25 similarity score, as well as the top-5 most similar chunks. The results show very similar performance between the upper-bound oracle and RAG using the top-1 document. However, a significant drop is observed when using chunks, highlighting a known caveat of RAG systems and their sensitivity to chunking strategies.

The solid black line represents the CPT performance, serving as the baseline method for injecting knowledge from unstructured text. The other colored lines correspond to each augmentation method under comparison. For all methods, we repeated the synthetic generation $N$ times (shown on the x-axis), using a sampling temperature of 1.0 to introduce variation in the generated examples.

For the RTW method, we report separate results for different prompt configurations: (i) using all four proposed prompts (easy, medium, hard, and QA-style), and (ii) two variations -- one excluding the QA-style prompt and another using only the QA-style prompt~\cite{maini-etal-2024-rephrasing}. This separation was implemented to measure the impact of the generated QA pairs and to mitigate the risk of indirect contamination, where the model might generate QA pairs similar to those encountered during testing, thus making the task artificially easier. Although we ensured that the QA pairs generated by RTW do not overlap with those in the test set, we report these results separately as a precaution.

The results indicate a monotonic increase in performance when using the RTW and Para methods, with RTW achieving a performance level comparable to document-level RAG, only a few points below the upper bound. The RTW variant that excludes the QA-style prompt and the Para method both yield similar performance, approaching that of chunk-based RAG. However, the superior performance of RTW suggests that leveraging multiple prompts to generate textual variations is more effective than relying on a single paraphrasing prompt.

The RTW variation that uses only QA-style prompts outperforms the configuration using only the other three prompts. This result indicates that synthetically generated QA pairs play an important role in increasing the model's ability to answer questions based on learned information. However, concerns about indirect contamination remain, suggesting that the no-QA variant may be better suited for real-world applications.

The IPT method scored lower than expected, hovering around the CPT baseline. This might be due to using a one-shot approach instead of the few-shot approach that worked best in the original study -- a strategy that was not feasible here because of the longer texts used, which would exceed the model’s context-length limit.

\subsection{Learning-forgetting tradeoff}

\begin{figure}
    \centering
    \includegraphics[width=\linewidth]{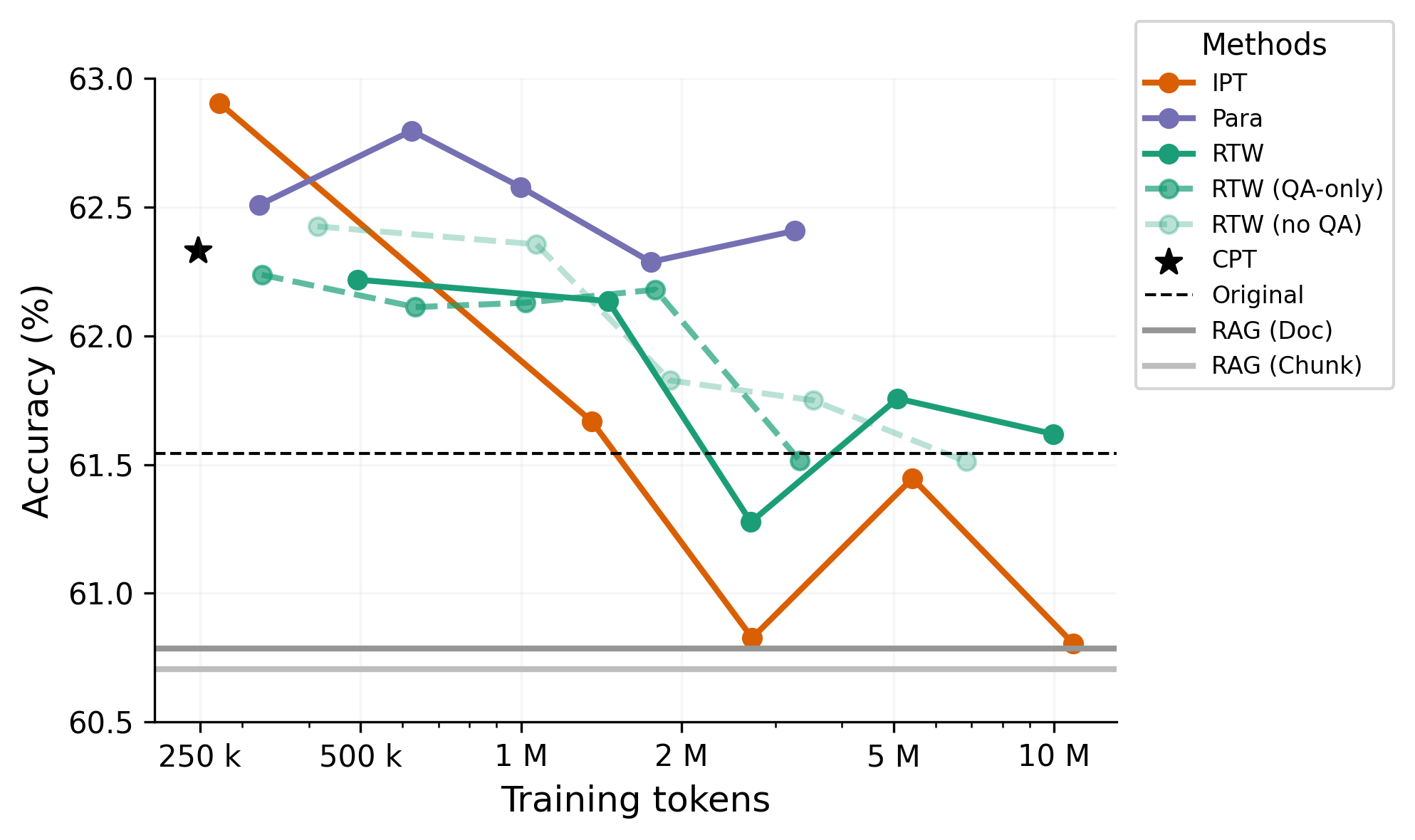}
    \caption{Comparison of average accuracy across seven control datasets against the number of training tokens. Each point on the lines represents a different variation level (1, 5, 10, 20, 40). The results indicate a trend of accuracy degradation in the control sets as more variations are introduced during training, suggesting the onset of catastrophic forgetting of previously learned capabilities. This effect is also observed in RAG variants, which show performance degradation when exposed to in-domain data and evaluated on out-of-domain data. Note that the x-axis is on a logarithmic scale.}
    \label{fig:control}
\end{figure}

To measure the possible catastrophic forgetting, we evaluated each checkpoint on the control dataset and averaged the accuracy across seven different tasks. This gives us a measure of how the model performs on tasks it previously knew.

Fig.~\ref{fig:control} compared the control dataset accuracy with the total training tokens of each method on each number of generated variations (1, 5, 10, 20, and 40). The CPT baseline is indicated by a star, since its amount of training tokens is fixed, and we compared with the original performance and the RAG variants.

To evaluate the RAG performance, we prepended the retrieved context on each evaluated prompt, following the same prompt used on non-RAG evaluations. Thus, their evaluations are exactly the same as the other methods, the only difference is that the retrieval result is included in the context. It is noteworthy that both RAG approaches lead to the highest degradation compared to the other methods, with the exception of IPT. This performance degradation highlights the caveats of using RAG-based approaches, where the retrieved context may \textit{confuse} the model on out-of-domain tasks.

Surprisingly, all continued pre-training methods evaluated result in an increase in performance when training with a small number of tokens. This overall performance improvement was observed across all datasets except BoolQ and WinoGrande. One possible explanation for this result is that we start from an instruction-tuned model and evaluate it by computing log probabilities on multiple-choice tasks using the Language Model Evaluation Harness framework~\cite{eval-harness}. This evaluation approach may favor base models over instruction-tuned models due to probability calibration. Thus, this short continued pre-training on the next-token prediction task may resemble the original pre-training objective, leading to performance gains on some control datasets. However, we leave a deeper investigation of this phenomenon to future work.

The results indicate that, as the amount of training tokens increase, the average performance tends to drop. This result is consistent with the catastrophic forgetting phenomenon. This is not entirely true for the Para method, which oscillated on similar performances, but a more informed conclusion would required more tokens to check if the trend holds, since it is the method that resulted on the lowest amount of tokens.

Despite the difference in the in-domain accuracy, i.e., the accuracy on the recent news dataset, both RTW methods achieved similar performance on the control dataset, even though the dataset with all four types of prompt resulted on a slightly higher amount of generated tokens.

The IPT method exhibited the largest drop in performance, suggesting that incorporating synthetic instructions generated by its synthesizer accelerated the model's forgetting of previously learned capabilities. Moreover, IPT showed low performance both in-domain and out-of-domain, indicating that the one-shot generation approach used in our experiments may be more detrimental than beneficial in our evaluated scenario.
% the same considerations should be taken into account for this result, just as in the previous section.

\subsection{Can models augment themselves?}

\begin{figure}
    \centering
    \includegraphics[width=1\linewidth]{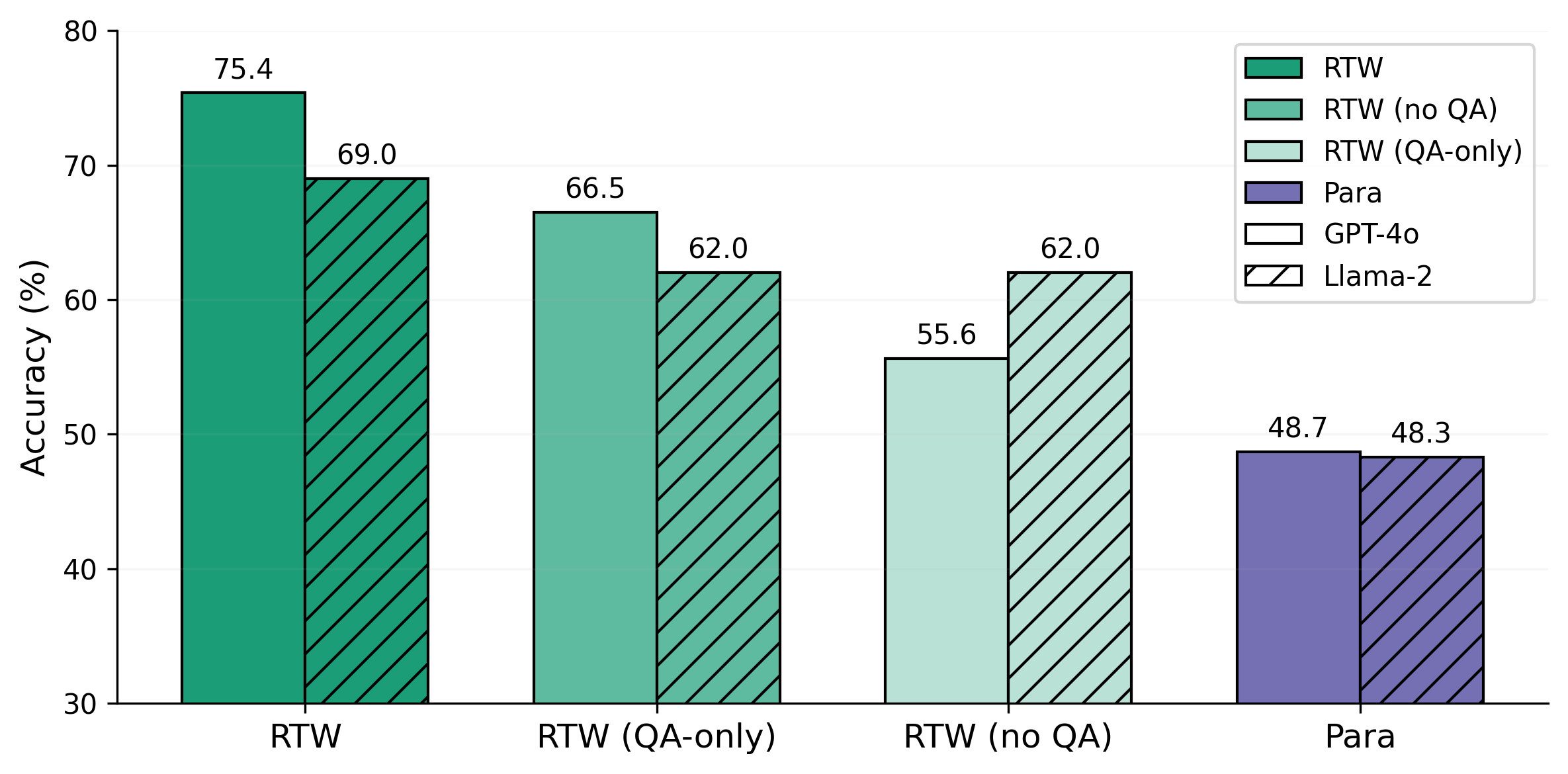}
    \caption{Performance on the TiEBe test set of recent events using different augmentation methods that leverage external models to generate synthetic examples. We present variants of the RTW algorithm and Para, utilizing GPT-4o and Llama-2-7B. Specifically, we highlight that the model used for training can also generate synthetic data to augment its own training.}
    \label{fig:generators}
\end{figure}

Previous results show that models can effectively learn new information by continuous pre-training on new data and benefits from synthetically augmented data. In this section, we investigate the role of the generator model used to augment the dataset. We used the RTW and Para techniques, since the IPT uses their specific synthesizer model, instead of a generic LLM.

Fig.~\ref{fig:generators} shows the comparison of the different generator models: GPT-4o and Llama-2-7B. For simply paraphrasing the content (Para), there was no significant difference of using a frontier model or the model itself to generate the synthetic training data. For the RTW, which uses varied prompts to augment the data, it is inconclusive whether one model performs better than the other because using all four prompts the GPT-4o lead to better results, but without the QA-style prompt, the model itself leads to better results.

% A discussion that can be made of it 
One potential caveat is that GPT-4o was the same model used to generate the questions for the original TiEBe dataset~\cite{almeida2025tiebebenchmarkassessingcurrent}. Thus, an unwanted indirect contamination may explain it leading to better performance, since the model might have generated similar questions for the knowledge injection methods and for our test set. This hypothesis also explains why removing the QA prompt from RTW resulted in a large drop in performance when using GPT-4o as a generator. However, this drop is smaller when using LLaMA-2-7B as the generator.

Even though the results of the QA-only variant are higher with the GPT-4o generator, there is no difference when using the model itself as the generator. This highlights the performance achieved by the no-QA variant, which avoids the risk of exposing the model to similar QA pairs in the test, while showing the generalization of the learned information due to solely training the model on rephrased versions of the original text.

Achieving comparable results with the smaller LLaMA-2-7B model and a state-of-the-art model is particularly noteworthy. It shows that the model can generate synthetic data to enhance its own capabilities, suggesting the possibility of continuous or iterative self-improvement through the ingestion of newly generated data.

% \begin{figure}
%     \centering
%     \includegraphics[width=1\linewidth]{images/plot-temperature.png}
%     \caption{Enter Caption}
%     \label{fig:temperature}
% \end{figure}

\section{Conclusion}

% In this study we compared different approaches to inject new knowledge into LLMs by continual training on unstructured text data. We compared different augmentation strategies with direct continual pre-training and RAG showing promising results in augmentation methods.

In this work, we investigated methods for injecting small-scale, unstructured knowledge into LLMs and examined the tradeoff between learning new facts and retaining prior knowledge. We found that simple continued pre-training yields modest improvements, while RAG can be effective but often degrades performance on unrelated tasks.

Our results highlight the importance of synthetic data augmentation: models trained on diverse rephrasings (e.g., RTW) learn new information more effectively while avoiding catastrophic forgetting. Notably, models can generate their own augmentation data, opening avenues for self-improving updates without external supervision.

Our findings emphasize the need for diverse training inputs to enhance knowledge acquisition while minimizing the degradation of previously learned information. We hope our released datasets and code will support future research on efficient knowledge injection. 

\bibliographystyle{IEEEtran}
\bibliography{main}

%\begin{table*}[t]
%  \centering
%\end{table*}

\end{document}